# A Novel BGCapsule Network for Text Classification

**Akhilesh Kumar Gangwar[1,2] and Vadlamani Ravi[1, *]**

[1]Center of Excellence in Analytics,
Institute of Development Research in Banking Technology
Castle Hills Road #1, Masab Tank, Hyderabad-500057 India
[2]School of Computer and Information Sciences, University of Hyderabad-500046, India
*gangwar.akhilesh1993@gmail.com; rav_padma@yahoo.com*

**Abstract**

Several text classification tasks such as sentiment analysis, news categorization, multi-label classification and opinion classification are challenging problems even for modern deep learning networks. Recently, Capsule Networks (CapsNets) are proposed for image classification. It has been shown that CapsNets have several advantages over Convolutional Neural Networks (CNNs), while their validity in the domain of text has been less explored. In this paper, we propose a novel hybrid architecture viz., BGCapsule, which is a Capsule model preceded by an ensemble of Bidirectional Gated Recurrent Units (BiGRU) for several text classification tasks. We employed an ensemble of Bidirectional GRUs for feature extraction layer preceding the primary capsule layer. The hybrid architecture, after performing basic pre-processing steps, consists of five layers: an *embedding layer* based on GloVe, *a BiGRU based ensemble layer, a primary capsule layer, a flatten layer and fully connected ReLU layer followed by a fully connected softmax layer*. In order to evaluate the effectiveness of BGCapsule, we conducted extensive experiments on five benchmark datasets (ranging from 10,000 records to 700,000 records) including *Movie Review* (*MR Imdb* 2005), *AG's News* dataset, *Dbpedia ontology* dataset, *Yelp Review Full* dataset and *Yelp review polarity* dataset. These benchmarks cover several text classification tasks such as news categorization, sentiment analysis, multiclass classification, multi-label classification and opinion classification. We found that our proposed architecture (BGCapsule) achieves better accuracy compared to the existing methods without the help of any external linguistic knowledge such as positive sentiment keywords and negative sentiment keywords. Further, BGCapsule converged faster compared to other extant techniques.

## 1 Introduction

Text classification is one of the most basic and important applications of machine learning. Traditionally, the use of term frequency inverse document frequency (tf-idf) in forming the document-term matrix as a representation of documents followed by invoking general classifiers such as naïve bayes, support vector machines (SVM) or logistic regression has been the de facto standard for text classification. Recently, however, proliferation of powerful neural embedding approaches have made it possible to find distributed representations of words and documents in an efficient manner [1] which further led to higher accuracies in text classification. The major deep learning models employed in text classification are largely based on convolutional neural networks (CNNs) and recurrent neural networks (RNNs).

Despite great success, these deep neural networks have some inadequacies. In the case of text data, these deep learning models heavily depend on the quality of instance representation of text data. Here instance could be a sentence, documents or paragraph. CNN, RNN based text classification requires huge amount of training data to learn and they do not perform so well on small datasets.

Meanwhile, in the image classification domain, capsule networks proposed by Sabour et al.[2], proved to be effective at understanding spatial relationships in high levels of data by employing a whole vector of instantiation parameters. We applied modified and extended version of this network structure to the classification of text, and argue that it has advantages in this field.

### 1.1 Motivation

Text classification using deep learning is based on the concept of feature extraction from text data. Text feature extraction can be accomplished using convolutional neural network (CNN) or recurrent neural network (RNN). CNN performs n-gram based feature extraction and RNN employs window-based feature extraction. If the length of a sentence is less than 10, then CNN and gated recurrent unit (GRU) yield comparable performance. While CNN captures local features, it does not capture sentiments from long sentences. Other problem with these deep learning models is that they require large amounts of training data for producing good results.

Further, CNN and RNN based methods use max pooling concept for feature routing which is called static routing. It tends to lose important context and information. Hence, it is not much useful for text datasets.

We proposed a new BGCapsule architecture in order to overcome these issues. BGCapsule performs better on less data also compared to CNN. Sabour et al.[2] proposed dynamic routing algorithm to pass features from lower layer to higher layer. Dynamic routing algorithms unlike static routing such as max pooling yield better performance.

### 1.2 Contributions

In this work, we build faster and robust text classification model based on the capsule network. The main contributions are as follows: -
- To the best of our knowledge, hybrid of the two Bidirectional gated recurrent unit (BiGRU) ensemble and Capsule is the first attempt for multi-label and multiclass classification. While BiGRU ensemble takes care of feature extraction, the capsule net takes care of classification.
- We demonstrate that the BGCapsule achieves state-of-the-art performance without any external knowledge of dataset.
- We tested the performance of BGCapsule on five benchmark datasets of different sizes and different tasks such as binary classification and multiclass classification with that of the extant methods. Our result shows that capsule model is competitive and robust. Our model performed well on small as well as on big datasets compared to other state-of-the-art methods.
- We performed ablation study by changing the composition of the hybrid as well where we compared the performance of BGCaspule with that of BiGRU + Max Pooling and CNN+Capsule Network.

The rest of the paper is organized as follows: section 2 presents related work; section 3 presents in detail our proposed model; section 4 describes the benchmark datasets on which we tested our model; section 5 presents discussion of the results and finally, section 6 concludes the paper and presents future research directions.

## 2      Related Work

Text classification tasks are also impacted by the deep learning revolution that is witnessed of late. Text classification using traditional machine learning mainly focuses on feature engineering. For achieving better performance, text classification heavily depends on the choice of feature representation of text because many representations reflect semantic meaning of neighboring words. Hence, it is better to capture the context as well.

Critical preprocessing phase of text classification is feature representation. The most prominent feature representation is the document term matrix, which, in turn, has a few variations like term count, term frequency, and tf-idf scores. [3]. Despite that, it is an open research challenge to obtain better feature representation for text corpora [4]. Neural network-based methods met with great success in natural language processing (NLP) tasks by offering simple and effective approaches to learn distributed representations of words and phrases [1], Many deep learning models have been applied to text classification, including Recursive Autoencoder [5,6] , Recursive Neural Tensor Network [6], Recurrent Neural Network [7], LSTM [8], and GRU [9].

In all these works, deep learning networks are used as feature extractors. For instance, CNN is used for n-gram extraction, where, the n-grams (features) are fed to multilayer perceptron to classify the text corpora. On the other hand, sequential networks like LSTM, GRU, BiLSTM etc. are also used for feature extraction. However, the advantage of these sequential networks is that the feature extracted using them capture some context information. These networks are window-based feature extractors.

Kim [10] has proposed first CNN based text classification model. He has shown how CNN is able to detect n-gram features for text classification-based sentiment analysis. He introduced three variants of CNN viz., random CNN, static CNN and non-static CNN. Zhang et.al.[11] developed character based embedding method for text classification. Character based embedding method helps in out-of-vocabulary cases (OOV). If pretrained embedding is not trained on particular words, we called it OOV case.  Conneau et al.[12] proposed the very deep convolutional neural network using skip connection. They used residual blocks for taking the advantage of large depth models. Miyato et.al [8] proposed semi supervised learning-based classifier. They used LSTM in adversarial manner. Kim et.al. [13] proposed capsule net for text classification. They used two types routing algorithms. Wang et al.[14] proposed hybrid capsule net for text classification. They have used only small datasets to check the performance of the network. Zhao et. al. [15] proposed capsule network with dynamic routing for text classification. They reported that capsule network yields significant improvement in accuracy when transfer single-label to multi-label text classification. Single-label corresponds to assigning a particular text document to one class out of N classes and multi-label corresponds to assigning a given sample to more than one class. Zhang et al. [16] proposed attention based capsule network for relation extraction. They mentioned that capsule network converts the multi-label classification problem into a multiple binary classification problem. Miyato et.al [17] proposed adversarial training methods for semi-supervised text classification. They mentioned that adversarial and virtual adversarial training have good regularization performance in sequence models on text classification tasks. Feng et.al [18] proposed image classification using capsule guided by external textual knowledge. Their proposed model performed better than existing capsule using additional text data.

Quite coincidentally, Du et. al [19] proposed a similar capsule based hybrid neural network for short text classification. They used capsule network with attention and CNN, RNN architectures. They performed experiments on two datasets including Movie Review (MR) dataset [10], which we also analyzed. Our proposed architecture is different from theirs in terms of feature extraction; while we used an ensemble of two BiGRUs of different unit sizes such that it captures different text features and concatenated it for the use of next layer, they used only one GRU layer followed by attention concept. While we validated our model's performance on 5 datasets including Movie Review (MR) dataset, they tested on only two datasets.

Our proposed model achieved better accuracy on MR dataset compared to that of Du et al. [19] despite not using attention layer. Largest data tested by us has 650,000 samples, whereas the largest dataset tested by them has 10000 samples. We performed 10-fold cross validation for the MR dataset, while they performed hold out method of testing.

## 3     Proposed Model - BGCapsule Network

The architecture of the proposed BGCapsule network, depicted in Figure 1, is a variant of the capsule network proposed [2]. It consists of six layers: embedding layer, BiGRU based ensemble layer, capsule network (which has layers too), flatten layer, fully connected ReLU layer and fully connected softmax layer. We elaborate the key components in detail as follows:

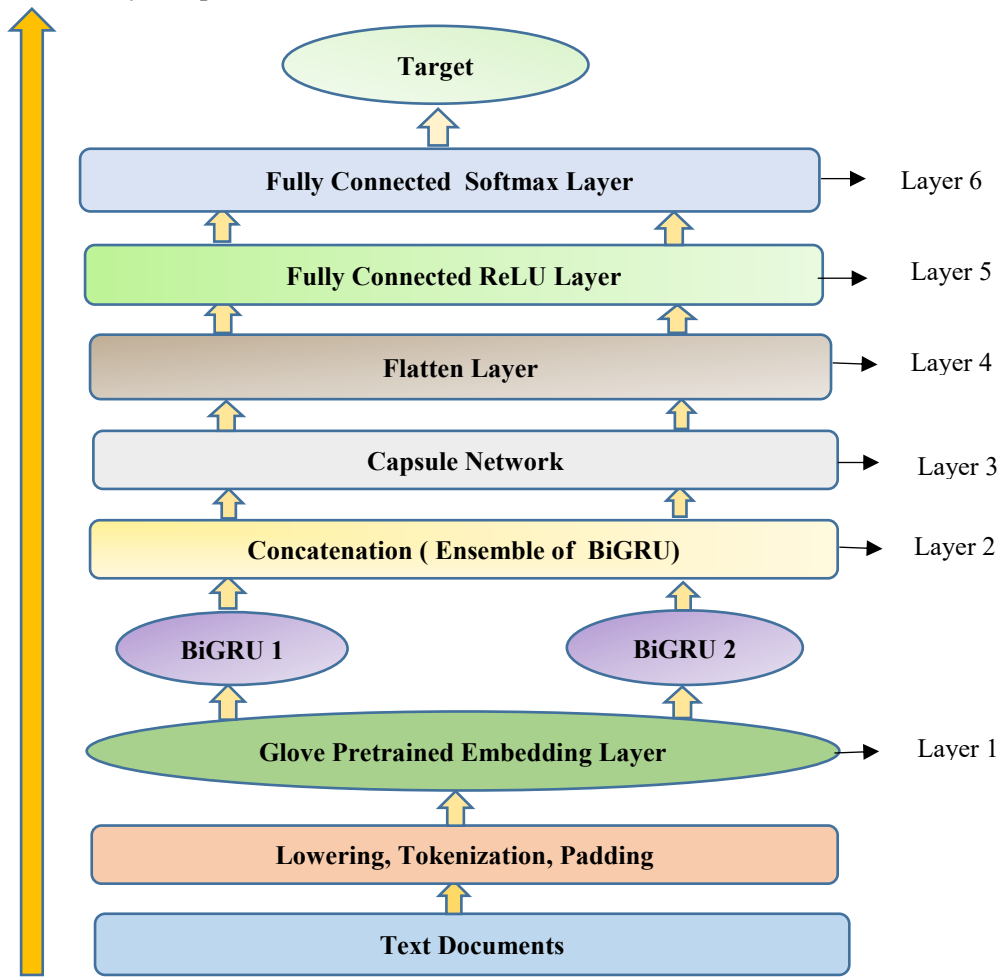

Figure 1: Proposed Architecture of BGCapsule Network for text

We performed lowering, tokenization and padding on the text documents as part of the preprocessing tasks. In lowering phase, we converted every sentence to lowercase. Then, we tokenized the sentences and assigned a particular integer index to each token. We then selected maximum length limit for each sentence. We performed pre-padding with zeros up to maximum length. Padded zeros represent that there is no word. It is meant for just making each document of same length. After that, we fed padded sentence to the embedding layer. Embedded layer then converts each token of a document to N dimensional vector and it converts each padded zero to N dimension zero vector.

3.1 Embedding Layer

Word embedding are obtained through distributed context vector model and dimensionality reduction. Distributed context vector model capture the context of the words in which they present in corpus. These vectors have dimensionality of the vocabulary size of the corpus [20]. These vectors obtained by training over the corpus and calculating the co-occurrence of a word. Mikolov, et al [1], proposed Word2Vec in two variants: (i) continuous bag of words and (ii) Skip-gram model. Both models capture the co-occurrence of one window at a time. Recently, in an open-source project at Stanford, Global Vectors for word representation (GloVe) [21] tries to capture the counts of overall statistics as to how often it appears.

In this paper, for each sentence, we use pre-trained GloVe word embedding [22]. It is an unsupervised technique for obtaining vector representation of words. Training is conducted on aggregated global word-word co-occurrence statistics from a corpus, and the representations thus obtained portray interesting linear substructures of the word vector space. We used GloVe model because it has the benefits of the Word2Vec's skip-gram model [23] for word analogy tasks, and matrix factorization methods which exploit global statistical information. Finally, each sentence is collapsed into a matrix M of size p x v.

$$x_i = \{w_1, w_2, \ldots \ldots w_p\} \in M^{p \times v} \quad \text{———————} \quad (1)$$

where, w1, w2….wp are the words of sentence padded up to a user-defined length, p; v is the length of word vector representation. We considered N number of such matrices, where N is the batch size representing the number of text documents. Thus, our data become three dimensional array of size $N \times v \times p$.

3.1    BiDirectional Gated Recurrent Unit Layer (BiGRU)

GRU [11] is a type of RNN. A Recurrent Neural Network (RNN) is a class of neural networks, which can handle temporal information with temporal inputs and outputs [24]. Conventional neural network has connection between the units in different layers, but in RNN it has connections between hidden units [24] forming directed cycle in same layer. Due to the recurrent connections it is able to transmit temporal(sequential) information. Therefore, RNN outperforms other networks extracting the temporal features. Bidirectional Recurrent Neural Network (BiRNN) [25] connects two hidden layers in opposite directions to the same output. The output layer can get information from past (backwards) and future (forward) states simultaneously.

BiRNN increases the amount of input information available to the network. BiGRU [26] is the most advanced RNN and is less complex compared to BiLSTM. BiGRU works as a better window-based feature extractor. LSTM architecture is very effective, but also complex. Due to the complexity LSTM is hard to analyze and it is slow as well. Gated recurrent unit (GRU) was recently introduced by Cho et al. [11] as an alternative to LSTM. It was subsequently shown by Chung et al. [26] that it performed comparably to the LSTM on several (non-textual) datasets. We make use of ensemble of two BiGRUs to deeply learn the semantic meaning of the sentences.

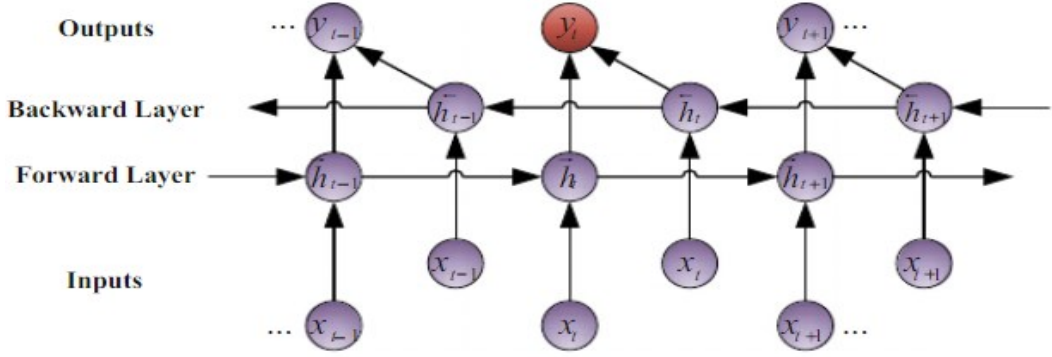

Figure 2: Bidirectional RNN architecture

Conventional RNN can only capture temporal information of on direction, however BiRNN can capture temporal information of both forward and backward direction. In figure 2, we have depicted BiRNN diagram, $x_{t-1}$, $x_t$, $x_{t+1}$ are temporal inputs and $y_{t-1}, y_t, y_{t+1}$ are temporal outputs. $\vec{h}_{t-1}, \vec{h}_t, \vec{h}_{t+1}$ are states for forward sequence and $\overleftarrow{h}_{t-1}, \overleftarrow{h}_t, \overleftarrow{h}_{t+1}$ are state of backward sequence.

We employed ensemble BiGRU layer for feature extraction. It extracts the features better than convolutional neural network (CNN). CNN only detects n-grams whereas BiGRU detects context and pattern also in sequence manner.

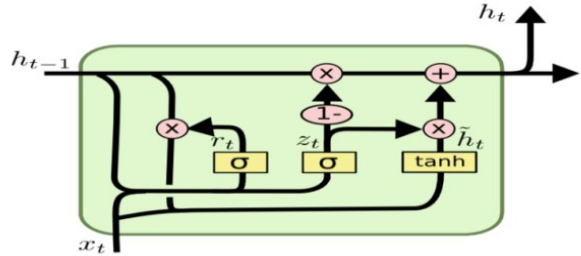

Figure 3: GRU cell architecture

GRU has two gates: reset gate and update gate. These are useful in handling long term dependency. Fig. 3 depicts the GRU cell architecture. Here, $h_t$, $h_{t-1}$ are the output of current state and previous state, $x_t$ is the input to current state, $[h_{t-1}, x_t]$ is the concatenation of $h_{t-1}$, $x_t$, Update gate $z_t$ and forget gate $r_t$ are obtained through the dot product of $W_z$ & $[h_{t-1}, x_t]$ and $W_r$ & $[h_{t-1}, x_t]$ respectively for time stamp t.

$\sigma$ and tanh are sigmoid and tanh layer respectively. By using $r_t$ and $z_t$ we calculate the output of the cell state $h_t$ for time stamp t.

$$h_t = (1 - z_t) \times h_{t-1} + z_t \times h_t \qquad (2)$$
$$h_t = \tanh(W.[r_t \times h_{t-1}, x_t]) \qquad (3)$$
$$r_t = \sigma(W_r.[h_{t-1}, x_t]) \qquad (4)$$
$$z_t = \sigma(W_z.[h_{t-1}, x_t]) \qquad (5)$$

Main property of GRU cell state, the horizontal line running through the top of the Figure 3, is that it can remove or add information to the cell state based on update and reset gate. With the help of the update and forget gates, we can handle the information passing from the previous state to the next state.

In the update gate, we get a vector which contains value between 0 to 1. This gate has point-wise multiplication operation. A sigmoid activation sqaushes values between 0 and 1. It helps to update or forget data because any vector getting point-wise multiplied by vector of 0 results in the values to disappear or be "forgotten." On the other hand, any vector multiplied point-wise by vector of 1 results in the same value. Therefore, that value stays the same or is "kept." The network can learn which data is not important and therefore can be forgotten or which data is important to be kept. The reset gate is another gate used to decide how much past information can be forgotten. GRU cell has fewer tensor calculation. Hence, it is faster than an LSTM cell. We used Bidirectional GRU network which is made of two GRU cells.

3.2     Capsule Network

Capsule network is proposed by Sabour et al.[2] for image classification task, which was demonstrated to be a better image classifier for learning spatial relationship. Our goal is to hybridize a capsule network with an ensemble of BiGRUs for text classification. Capsules have the ability to represent attributes of partial entities, and express semantic meanings in a wider space by expressing the entities with a vector rather than a scalar. In this regard, capsules are suitable to express a sentence or document as a vector. Fig. 1 depicts the general structure of the proposed model. The architecture of the capsule network, depicted in Fig. 4, is described as follows:

*Primary Capsule Layer*: This is the first capsule layer in which the capsules replace the scalar-output feature detectors (CNN or BiGRU) with vector-output capsules to preserve the instantiated parameters for each feature, such as the local order of words and semantic representations of words.

*Connection Between Capsule Layers:* Capsule network generates the capsules in next layer using the principle of "routing-by-agreement". This process predominantly replaces the pooling operation. Pooling operation loses some important information like angle, position and cannot capture equivarance. Equivarance means the internal representation capture the properties of object, meaning that if we change the internal representation then it also changes the object. If there is no information lost, then it helps make robust prediction otherwise fooling a network with pooling operation becomes easy [27]. In fig 4, $u_i$ is output of ith capsule of lower layer L and $v_j$ is output of jth capsule of next layer L+1.

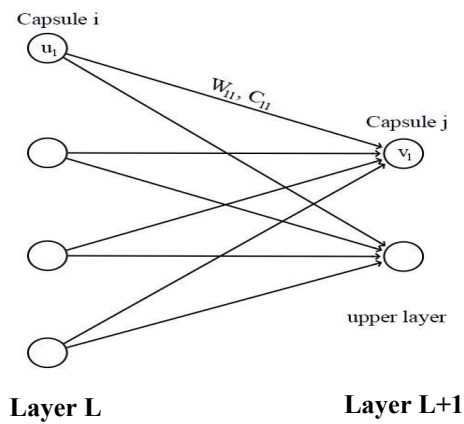

**Layer L**                **Layer L+1**

Fig. 4: Capsule layers connections

Between two neighboring layers L and (L+1), a "prediction vector" uj|i is first computed from the lower layer L capsule output ui , by multiplying it with a weight matrix Wij. In the capsule network, ui  is the

input vector and vj is the output vector for jth capsule of L+1 layer. For getting output of next layer capsule, apply transformation $W_{ij}$ to low level layer capsule output.

$$u_{j|i} = W_{ij} u_i \qquad (6)$$

Then in (L+1) parent layer, a capsule sj is generated by linear combination of all the prediction vectors with weights cij.

$$s_j = \sum_i c_{ij} \cdot u_{j|i} \qquad (7)$$

$c_{ij}$ represents the coupling coefficients. Dynamic routing algorithm is required to calculate the value of coupling coefficient and $\sum_i c_{ij}$ are designed to sum to one. Coupling coefficient $c_{ij}$ reflects the effectiveness of capsule i to activate capsule j.

For maintaining nonlinearity, instead of applying a sigmoid, tanh or ReLU [28] activation function, capsule network use squashing function to sj. It transforms the activity vector (next layer L+1 capsule output vector) vj to length between 0 and 1. Equations 9 and 10 show that squashing function shrinks small vectors to almost zero and large vectors to unit vectors. When the value of $s_j$ is very small (large), then the value given by equation 8 tends to zero (one). Hence, we can approximate equation 8 by equation 9 (10). This squashing function limits the length of capsule with non-linearity. By this process, the short vectors are pushed to shrink to zero length, and long ones are pushed to one.

$$v_j = \frac{|s_j|^2}{1+|s_j|^2} \frac{s_j}{|s_j|} \qquad (8)$$

$$v_j \approx |s_j| s_j \quad , \text{when } s_j \text{ is very small} \qquad (9)$$

$$v_j \approx \frac{|s_j|}{s_j} \quad , \text{when } s_j \text{ is large} \qquad (10)$$

*Dynamic Routing:* The capsule network updates the weights of the coupling coefficients through an iterative dynamic routing process [2] and determined the degree to which lower capsules were directed to upper capsules. The coupling coefficient is determined by the degree of similarity between the standard-upper and prediction-upper capsules.

Recall that the prediction vector uj|i and activity vector vj (output of jth capsule of next layer L+1) are already computed. Prediction vector uj|i represents votes from the capsule i for the output capsule j above. If the input vector is highly similar to the voted vector, we conclude that both capsules are highly related. For similarity measure we use dot product between the prediction and the activity vector.

$$b_{i|j} \leftarrow u_{j|i} \cdot v_j \qquad (11)$$

Due to the dot product bij not only takes into account likeliness but also feature properties. The value of, bij will not be high, if the activation ui of capsule i is low since uj|i length is proportional to ui. i.e. bij should remain low between the lower layer L capsule and the parent layer capsule if the lower layer capsule is not activated. The value of cij is obtained by using softmax of bij:

$$C_{ij} = \frac{exp(b_{ij})}{\sum_k exp(b_{ik})} \qquad (12)$$

Algorithm updates the value of bij iteratively in multiple iterations to make it more accurate.

$$b_{ij} \leftarrow b_{ij} + u_{j|i} \cdot v_j \qquad (13)$$

In this paper, we have used dynamic routing algorithm proposed by Sabour et al.[2] for features routing. Dynamic routing is used to update the parameters like coupling coefficient $c_{ij}$ which expresses the connection between a capsule and its parent capsule. As routing algorithm is still not enough to update all parameters, backpropagation is used to calculate the weight matrix Wij. The value of $c_{ij}$ has to be re-initialized before dynamic routing calculation begins.

---

Algorithm 1 Dynamic Routing Algorithm

---

1. procedure ROUTING (uj|i , r , L)
2.   for all capsule i in layer L and capsule j in layer (L +1): $b_{ij} \leftarrow 0$.
3.   for r iterations do
4.     for all capsules i in layer L: ci $\leftarrow$ softmax(bi)
5.     for all capsule j in layer (L+1): $s_j \leftarrow \sum_i c_{ij} \cdot u_{j|i}$
6.     for all capsule j in layer (L+1): $v_j \leftarrow squash(s_j)$
7.     for all capsule i in layer L and capsule j in layer (L+1): $b_{ij} \leftarrow b_{ij} + u_{j|i} \cdot v_j$
    return vj

---

### 3.3 Capsule Flattening Layer

The capsules in this layer are flattened into a list of capsules. If we have N number of capsules each with dimension D then the output of Flatten layer is a vector of dimension $N * D$. Output of the flatten layer represents important features. These are passed on to the multilayer perceptron (MLP) layer.

### 3.4 Fully Connected ReLU / Softmax Layer

Here, we used 2 fully connected layers with ReLU activation[28] for classification the features. Last fully connected layer with softmax function for obtaining the probabilities in for multiclass classification.

### 4     Description of the Datasets Performance Metrics

To validate the effectiveness of our BGCapsule network, we performed experiments on five benchmark datasets. Out of the five datasets, AG's News, DBPedia, Yelp Review Polarity and Yelp review Full are introduced by Zhang et al.[11]. The fifth one dataset is Movie Review (MR) taken from [10] . These benchmark datasets cover several text classification tasks such as sentiment analysis, ontology classification, news categorization and opinion classification. The details are presented in Table 1.

Movie Review (MR): MR (Imdb 2005) is a movie review dataset which contains reviews of movies in English [12]. It contains 5331 positive and 5331 negative reviews. We performed 10-fold cross validation (10FCV) on this dataset. This dataset is used for sentiment analysis task.

AG's news corpus: This is English news categorization dataset introduced by Zhang et al. [13]. It is a multiclass dataset with 4 classes and 496,835 categorized news articles. It has 30,000 training samples and 1900 test samples for each class. It is a balanced dataset with three fields: title, description and target class.

DBPedia ontology dataset: This is ontology classification dataset introduced by Zhang et al.[13]. It is a multiclass dataset with 14 non-overlapping classes. It is constructed by DBPedia in 2014. It has 560,000 training and 70,000 test samples. Each class has 40,000 training samples and 5,000 test samples.

Table 1 Statistics of the datasets analyzed

| Dataset | Classes | Train Samples | Test Samples | Classification Task |
|---|---|---|---|---|
| MR (Imdb 2005) | 2 | 9596 | 1066 | Sentiment Analysis |
| AG's News | 4 | 120,000 | 7,600 | English news categorization |
| DBPedia | 14 | 560,000 | 70,000 | Ontology classification |
| Yelp Review Polarity | 2 | 560,000 | 38,000 | Sentiment Analysis |
| Yelp Review Full | 5 | 650,000 | 50,000 | Sentiment Analysis |

.

Yelp Review Polarity: This is a sentiment analysis dataset with binary classification. It is obtained from the Yelp Dataset Challenge in 2015 [13]. This dataset is converted to polarity dataset based on rating in yelp reviews. The polarity of a review is measured from rating. Based on rating, it is constructed to have two labels namely positive polarity and negative polarity. It has 560,000 training samples and 38,000 test samples.

Yelp Review Polarity: Yelp Review Full dataset is also obtained from the Yelp Dataset Challenge in 2015. It is a multiclass sentiment analysis dataset with 5 classes. It has 650,000 training samples and 50,000 test samples. Each class has 13,000 training samples and 10,000 test samples.

Performance Metrics: We have used only accuracy as the performance metric for five benchmark datasets namely Movie Review (MR) [10], AG's News [11], DBPedia [11], Yelp Review Polarity [11], Yelp Review Full [11], because other metrics like precision, recall, F1-score are not reported for the baseline state-of-the-art, Character-level CNN [11]and VDCNN [12].

4.2    Experimental Details

For data preprocessing, all the datasets are tokenized, and all words are converted to lowercase. We took 200-word limit for each document. If any document has less than 200 words, we used pre-padding by zeroes, after converting the rest of the words into corresponding index integer numbers. Thereafter, all the words, represented by index numbers are transformed by pre-trained embedding method GloVe. We used GloVe [21] pretrained model. The GloVe model trained on 2.2 million vocabularies, 840 billion tokens of web data from Common Crawl. This Glove embedding projected each word to a 300-dimensional vector. Each index integer number represents a token which is mapped to 300-dimension vector using pretrained word embedding. Zero index represent padded token which mapped to 300-dimension zero vector. The dimensions of BiGRU1 and BiGRU2 are 256 and 200 respectively.

The following are the hyper-parameters: we used recurrent dropout value of 0.25, SeLU activation function for multilayer perceptron and ADAM optimizer [23] and a batch size of 1000. We ran the experiments to 20 to 25 epochs. All code is implemented in Keras and Tensorflow.

# 5   Discussion and Results

The hybrid model is developed and executed on a workstation with 16GB NVidia Quadro P5000 GPU with 20 microprocessors having 128 CUDA cores each. The system configuration is Intel Xeon(R) CPU E5-2640 v4, 2.4GHz, 32 GB RAM, 40 core intel- i7 and 40 cores in Ubuntu 16.04 environment.

## 5.1 Classification Accuracies:

We compared the BGCapsule with two state-of-the-art methods, which used supervised text classification methods using five benchmark datasets. We report the Character-level CNN [11] and VDCNN [12] as baseline. The results, in terms of accuracy, for all datasets are presented in Table 3, where the best results are marked bold.

## 5.2 Dynamic Routing (Capsule Net) over Max Pooling

We have used a capsule layer for dynamic routing in the place of max pooling. The disadvantage of max Pooling is that it loses information. Max Pooling and other pooling like average pooling and k-max pooling are static routing algorithms because they have a rule of thumb that determines which features are routed to upper layer. In static routing, information loss occurs. On the other hand, Capsule net uses dynamic routing, where it takes weighted average of extracted features instead of selecting the best features as is done in max pooling. Thus, we have a better feature routing algorithm in capsule net in addition to a better feature extractor resulting in better results for our hybrid model compared to the state-of-the-art models.

## 5.3 Ablation Study

Ablation study [29] is useful for comparative study of architectures. Wherein we tinkered with the proposed architecture for performing ablation study [15]. It resulted in two more architectures namely BiGRU + Max pooling, CNN-Capsule net. In order to demonstrate the effect of capsule net and dynamic routing better, we removed the dynamic routing and primary capsule layer. We performed the experiments on all benchmark datasets with different ablation architectures. We trained and tested all benchmark datasets with BiGRU + Max pooling, CNN-Capsule net and compared the results with that of our proposed novel architecture BGCapsule. We presented ablation results in Table 2. From Table 2, we can see that BiGRU + Max pooling and CNN-Capsule network are outperformed by BGCasule in terms of accuracy.

BiGRU+ Max pooling: We used BiGRU for feature extraction followed by max pooling for routing the features. In max pooling, we selected dominating feature out of four features and drop the other three features.

CNN-Capsule Network: We have used CNN as feature extractor in place of BiGRU ensemble.

Number of trainable parameters in BGCapsule Network is more compared to max pooling does not have any parameters to decide how to route the features. In capsule network, dynamic routing will select features dynamically and uses weight for each feature. For important features it assigns larger weights. Therefore, dynamic routing is better compared to max pooling because the latter drops some non-important features and hence there is information loss in max pooling operation.

We can conclude from the ablation study in Table 2 that BGCapsle is the best compared to other two feature extractor and routing algorithms in terms of the accuracy.

Table 2: Test Set Accuracy in the Ablation study

| Dataset | BiGRU + Max Pooling | CNN+ Capsule Network | Proposed Model (BGCapsule) |
|---|---|---|---|
| AG's News | 91.86 | 92.02 | 92.59* |
| Dbpedia | 98.45 | 98.62 | 99.02* |
| Yelp Review Polarity | 94.67 | 95.14 | 96.62* |
| Yelp Review Full | 60.02 | 61.29 | 66.93* |
| Imdb(2005) | 81.2 (Mean of 10 FCV) | 81.9 (Mean of 10 FCV) | 84.90 (best fold) 82.93 (Mean of 10 FCV) |

Table 3: Test Set Accuracy of BGCapsule compared to the State-of-the-art

| Dataset | (CNN Sentence) [13] | (VDCNN Resnet) [14] | Proposed Model (BGCapsule) |
|---|---|---|---|
| AG's News | 92.36 | 91.33 | 92.59* |
| Dbpedia | 98.69 | 98.71 | 99.02* |
| Yelp Review Polarity | 95.64 | 95.72 | 96.62* |
| Yelp Review Full | 62.05 | 64.72 | 66.93* |
| Imdb (2005) | CNN[12]  CapsNet[15]  81.4           81.00 | VA LSTM*[19] 83.4 | 84.90 (best fold) 82.93 (Mean of 10FCV) |

* VA LSTM = Virtual Adversarial LSTM

We bettered the performance of all previous studies on all datasets. We obtained better accuracy with better margin in yelp review full dataset for sentiment analysis task. We obtained 4.87% higher accuracy than the one reported by Zhang et.al [13].

We can see from Table 3 that our proposed BGCapsule network achieves better accuracy than state-of-the-art models, without using any linguistic knowledge such as knowledge of positive and negative words. The IMDB and AG's News datasets are small compared to other datasets. Therefore, VDCNN [14] having 49 CNN layers was unable to outperform normal char CNN [12] model. However, our proposed BGCapsule is able to outperform both these models and achieved better accuracy because capsule network can capture the feature information correctly. Hence, it does not require lots of data to learn unlike CNN. Our model also performed well on large data like Yelp Review dataset, where we obtained 4.87% higher accuracy compared to Zhang et.al. [13]. Thus, we can infer that our model can perform better on small as well as large datasets.

We used BiGRU ensemble layer over CNN layer for feature extraction. Both CNN and RNN are good feature extractors. CNN is n-gram detector, but GRU will capture long range semantic dependency. We find that GRU performs better than CNN in cases, where we have to categorize text data based on entire sequence or a long-range semantic dependency rather than on some local phrases [12]. For example, if any long phrase contains negative word like "not","bad" but whole sentence have positive sentiment then in this case GRU will perform better. We opted BiGRU because it is a window-based feature extractor and extracts context successfully. Consequently, in order to extract better features, we employed an ensemble of two BiGRUs.

The reasons for the superior performance of the proposed model are as follows:

BGCapsule used the ensemble of two BiGRU network. We used two BiGRU with different unit size such that it captures different text features and concatenate it for the use of next layer. Concatenation of two BiGRU makes it a better feature extractor. Choice of number of BiGRUs that can be used in the ensemble and their unit size depends on RAM and it affects the computation. Therefore, we started with two BiGRUs for ensemble.

In the experiments, dimension of the capsule is an important hyper parameter. If the dimension of the capsule is large, then it can contain more feature information. However, larger dimension of capsule increases the computational complexity. We tried different dimensions of the capsule. For large dimensions of capsule, we observed that loss decreases slowly. Hence, we set the capsule's dimension to 20.

We obtained better results than the state-of-the-at studies for all datasets. Our model can perform very well on multi-label, multiclass, topic classification and sentiment analysis successfully. In YELP review full dataset our model outperformed character based convolutional neural net [12] and VDCNN because text feature extraction of BGCapsule is carried out by BiGRU layer.

## 6    Conclusions & Future Work

In this paper, we proposed a new hybrid architecture comprising an ensemble of Bidirectional Gated Recurrent units and Capsule Network for text classification domain. We used dynamic routing in capsule network and two BiGRU ensemble for feature extraction in place of CNN. We compared the proposed model with character based convolutional neural network and Very Deep Convolutional neural network. We observed that our proposed architecture BGCapsule is indeed useful for text classification based on five popular benchmark datasets and it has achieved best performance compared to the state-of-the-art.

In future, we would like to investigate the effect of routing algorithms on various tasks in multi-task learning by exploiting the potential of BGCapsule network. We will explore SOM and K-means based routing algorithms too. We also would like to investigate its usefulness in classifying numerical and image datasets.